\documentclass[11pt, a4paper, copyright, goog]{google}

\usepackage[numbers, sort&compress]{natbib}
\usepackage{bbm}
\usepackage{multirow}
\usepackage{ifthen}
\usepackage{makecell}
\usepackage{nicefrac}
\usepackage[most]{tcolorbox}
\usepackage{ragged2e}
\usepackage{setspace}
\usepackage{subcaption}
\usepackage{float}
\usepackage{listings}

\newboolean{ANONYMIZE}
\setboolean{ANONYMIZE}{false}

\usepackage{titlesec}
\titlespacing*{\section}{0pc}{2.76ex plus 3.7pt minus 2.8pt}{4.6pt}
\titlespacing*{\subsection}{0pc}{2.3ex plus 2.76pt minus 1.84pt}{1.84pt}
\titlespacing*{\subsubsection}{0pc}{1.84ex plus 2.3pt minus 1.4pt}{1.84pt}

\paperurl{}
\renewcommand{\today}{}

\uselogo{}

\title{MaxKernel: Agentic Kernel Generation for TPUs}

\correspondingauthor{sethusankaran@google.com}

\author[1,*]{Shangkun Wang}
\author[1,*]{Nina Cai}
\author[1]{Charles Hong}
\author[2]{Julian Walker}
\author[1]{Gerson Kroiz}
\author[1]{George Vanica}
\author[1]{Deepak Patil}
\author[1]{Andi Gavrilescu}
\author[1]{Hassan Sipra}
\author[1]{Sethu Sankaran}

\affil[*]{Equal contributions}
\affil[2]{Google DeepMind}
\affil[1]{Google}

\begin{abstract}
\vspace{-1.2em}
{\normalfont\mdseries\noindent Code: \url{https://github.com/AI-Hypercomputer/accelerator-agents/tree/main/MaxKernel}}\\[1.2em]
Designing and authoring high-performance custom kernels for accelerators is a complex task that requires deep hardware-level expertise. Large Language Models (LLM) can be leveraged together with real-time compiler feedback to build agentic systems for kernel generation. In this work, we present MaxKernel, a multi-agent system that implements three distinct paradigms for TPU kernel development: (1) a Human-in-the-Loop (HITL) agent for collaborative, step-by-step design; (2) an Autonomous (Auto) agent that executes a fully automated, metric/trace-driven optimization loop; and (3) a Graph-Based Autonomous Search that scales the Auto agent for global exploration of the design space. All three paradigms leverage a shared pool of specialized sub-agents to handle planning, implementation, self-debugging, testing, and hardware profiling. We evaluate MaxKernel on JaxBench, a comprehensive suite of 50 diverse kernel tasks for TPUs, alongside complex, real-world workloads from state-of-the-art open-source models. We demonstrate that MaxKernel consistently generates highly optimized implementations, matching expert hand-tuned baselines and delivering significant performance across the benchmark. Our agent is open-sourced and available \href{https://github.com/AI-Hypercomputer/accelerator-agents/tree/main/MaxKernel}{here}.
\end{abstract}

\begin{document}
\maketitle

\section{Introduction}
The continuous scaling of deep learning architectures has driven a growing demand for specialized hardware accelerators. To fully utilize the compute and memory bandwidth of these devices, engineers frequently bypass standard compiler pipelines to develop custom, optimized kernels. However, writing these custom kernels, whether in CUDA \cite{nickolls2008scalable, lindholm2008nvidia} and Triton \cite{tillet2019triton} for Graphics Processing Units (GPUs), JAX/Pallas \cite{bradbury2018jax} for  Tensor Processing Units (TPUs), Neuron Kernel Interface (NKI) for ~\href{https://awsdocs-neuron.readthedocs-hosted.com/en/latest/nki/index.html}{AWS Tranium}, Triton and Cude DSL for \href{https://ai.meta.com/blog/meta-mtia-scale-ai-chips-for-billions/}{Meta Training and Inference Accelerator (MTIA)} is a complex process. It requires extensive hardware expertise, specifically the ability to manually manage memory hierarchies (e.g., HBM versus SRAM/VMEM), orchestrate Direct Memory Access (DMA) pipelining, and derive multi-dimensional tiling strategies.

\begin{figure}[htbp]
    \centering
    \includegraphics[width=0.8\textwidth, keepaspectratio]{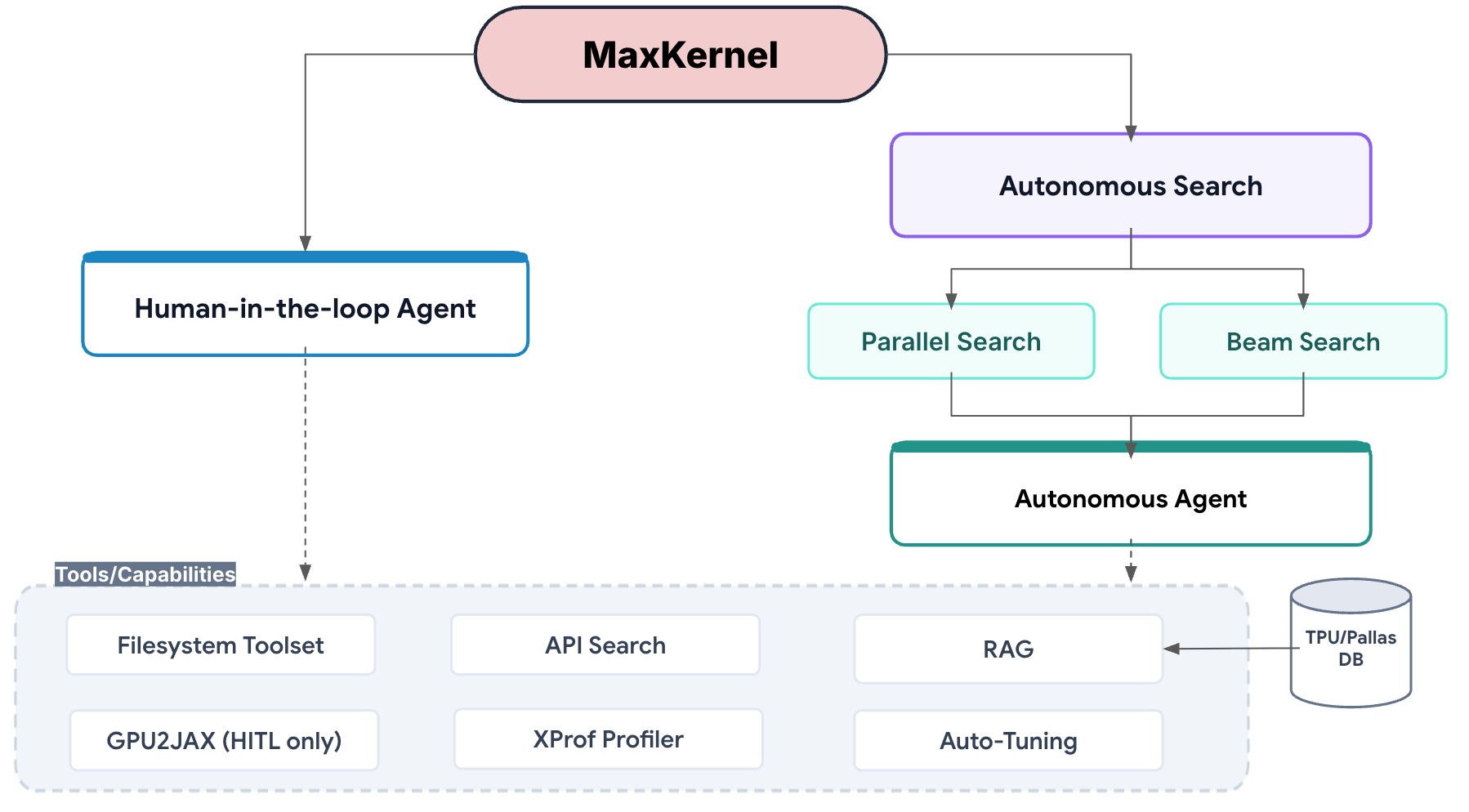}
    \caption{A schematic of the MaxKernel framework outlining the various modalities of operation and tools.}
    \label{fig:maxkernel_schem}
\end{figure}

Several works have shown that kernel generation benefits from test time scaling and agentic techniques  across GPUs, TPUs~\cite{hong2025autocomp,tschand2026jaxbench}, Tranium, NKI~\cite{tang2026nki}, MTIA~\cite{liao2026kernelevolve}, and other accelerators~\cite{hong2025autocomp, bryngelson2026apple}. Standard zero-shot or few-shot LLM approaches need to be augmented with real-time compiler feedback due to the rigidity of accelerator APIs, strict memory constraints, and opaque low-level compiler errors~\cite{tschand2026jaxbench}. Furthermore, even when an LLM generates functionally correct code, achieving optimal hardware performance typically requires an iterative process of empirical profiling, often utilizing tools like XProf \cite{hundt2026xprof}, configuration tuning, frequently changing APIs and context~\cite{hong2025autocomp,tschand2026jaxbench}, and optimization techniques including evolutionary~\cite{novikov2025alphaevolve},greedy search~\cite{pepe2026agentic,zhang2026memsearch}, and collaborative search~\cite{chen2026beyond}.

Traditional compiler optimization techniques, such as Halide \cite{ragan2013halide}, TVM \cite{chen2018tvm} and Ansor \cite{zheng2020ansor}, address these complexities by decoupling algorithm definitions from scheduling, yet they still require significant domain knowledge or expensive evolutionary search phases. Recent works, such as AutoComp \cite{hong2025autocomp}, have demonstrated the potential of using LLMs to navigate these search spaces for tensor programs, which may critical performance gains by abandoning complex ideas before the code can compile.

To address these challenges, we present \textbf{MaxKernel}, a multi-agent framework designed to automate and accelerate TPU kernel engineering. We benchmark our generated kernels on JAXBench \cite{tschand2026jaxbench} in addition to kernels for state-of-the-art OSS models, evaluating both functional correctness and execution time. While our primary evaluation focuses on TPU development via JAX/Pallas, the underlying agentic principles apply broadly to accelerator programming. MaxKernel  supports three distinct orchestration paradigms:
\begin{enumerate}
    \item \textbf{Human-in-the-Loop (HITL):} An interactive modality that keeps the human developer in control at critical decision points, ensuring safety and architectural alignment for complex kernels.
    \item \textbf{Autonomous Loop:} An automated, closed-loop agent that continuously iterates through planning, generation, testing, and trace-based hardware profiling to systematically optimize performance.
    \item \textbf{Graph-Based Autonomous Search:} An extension of the autonomous loop that models the kernel design space as a formal search graph. By leveraging different search algorithms (e.g., simple parallel search, beam search, etc), this paradigm conducts a broader exploration to navigate performance trade-offs and avoid local optima.
\end{enumerate}

In summary, our core contributions are as follows.
\begin{enumerate}
    \item \textbf{Modular Multi-Agent Architecture:} We introduce MaxKernel, a flexible, multi-agent framework comprising specialized sub-agents capable of writing, debugging, and tuning hardware-specific kernels. This architecture seamlessly integrates with compiler pipelines and empirical profiling tools (e.g., XProf) to provide critical closed-loop hardware feedback to the LLMs.
    \item \textbf{Versatile Orchestration Paradigms:} We detail three distinct execution workflows: Human-in-the-Loop, Autonomous Loop Pipeline, and Graph-Based Autonomous Search. This enables developers to dynamically trade off between manual architectural control, rapid automated iteration, and broad formal design-space exploration. 
    \item \textbf{Rigorous Empirical Evaluation:} We evaluate our framework on TPU optimization workloads using the JAXBench benchmark suite and SOTA OSS kernels. MaxKernel successfully bridges the gap between static LLM code generation and high-performance kernel engineering, achieving a 1.62× geometric mean speedup across the JAXBench workloads. Furthermore, on the eight production kernels where human-written kernels exist, MaxKernel achieves a 2.32× geometric mean speedup over the reference code, outperforming human-written ones (2.02×).
\end{enumerate}

\section{System Architecture \& Agent Paradigms}
MaxKernel is a modular, agent-based system that tackles complex kernel optimization by breaking it into focused tasks. A schematic is provided in Figure~\ref{fig:maxkernel_schem}. At its core, the architecture relies on a suite of specialized sub-agents dedicated to planning, implementation, testing, autotuning, and profiling. To accommodate varying levels of user control and task complexity, MaxKernel operates under three distinct execution paradigms: an interactive Human-in-the-Loop (HITL) modality for developer-guided exploration, an Autonomous Loop (Auto) for closed-loop iterative optimization, and a highly scalable Graph-Based Autonomous Search for systematic, parallel exploration of the kernel design space.

\subsection{Specialized Subagents}\label{sec:subagents}
The foundation of MaxKernel lies in its collection of task-specific sub-agents. By decomposing the complex process of kernel generation into manageable, specialized tasks, the system achieves higher reliability and allows for flexible orchestration. The key sub-agents include:

\begin{itemize}
    \item \textbf{Kernel Generation and Fix Agents}: Responsible for the core synthesis logic. This module is subdivided into a Planning Agent, which creates a high-level algorithmic optimization plan based on hardware specifications and the reference implementation, and an Implementation Agent, which translates the plan into valid Pallas code. To ensure syntactic and structural correctness, the validation and fix loop operates iteratively. The loop attempts to compile the generated code, interprets compiler feedback, and resolves errors before advancing to execution.
    \item \textbf{Testing and Verification Agents}: Ensure the numerical correctness of the generated kernels. A Test Synthesis Agent constructs comprehensive test suites, while an Execution Agent runs these tests securely on the target hardware (e.g., TPU). This module validates the optimized kernel against the reference implementation, strictly enforcing dynamically specified numerical tolerances for a given TPU generation.
    \item \textbf{Autotuning Agent}: Dedicated to hyperparameter optimization, this agent systematically explores the configuration space (e.g., block sizes, tile dimensions, etc) to discover the most performant hardware-specific parameters for a given kernel structure.
    \item \textbf{Profiling Agent}: Focused on performance extraction and bottleneck identification, the Profiling Agent captures and analyzes low-level trace data. By extracting key empirical metrics—such as execution latency, memory bandwidth utilization, and compute density—it provides actionable feedback to inform subsequent algorithmic optimization iterations. We leverage XProf for this purpose.
\end{itemize}

\subsection{Knowledge Store}\label{sec:knowledge}

Incorporating hardware-specific information, low-level framework documentation and technical reports is critical for enabling agents to effectively generate, debug, profile, and iteratively refine code. However, loading the comprehensive specifications into the context window is computationally expensive. To address this limitation, we employ a Retrieval-Augmented Generation (RAG) ~\cite{lewis2020retrieval} pipeline to dynamically surface relevant information from an external knowledge store during various execution phases. In this work, our RAG knowledge base is restricted to static sources, specifically comprising framework documentation, memory layout guides, and performance handbooks (e.g., Pallas, Mosaic, and XLA). Additionally, we explicitly exclude hand-tuned kernel code from the retrieval corpus; a primary objective of this study is to evaluate the extent to which agents can generalize and discover TPU-specific optimizations without relying on human-engineered baselines.

\subsection{The Human-in-the-Loop (HITL) Agent}

\begin{figure}[htbp]
    \centering
    \includegraphics[width=0.8\textwidth, keepaspectratio]{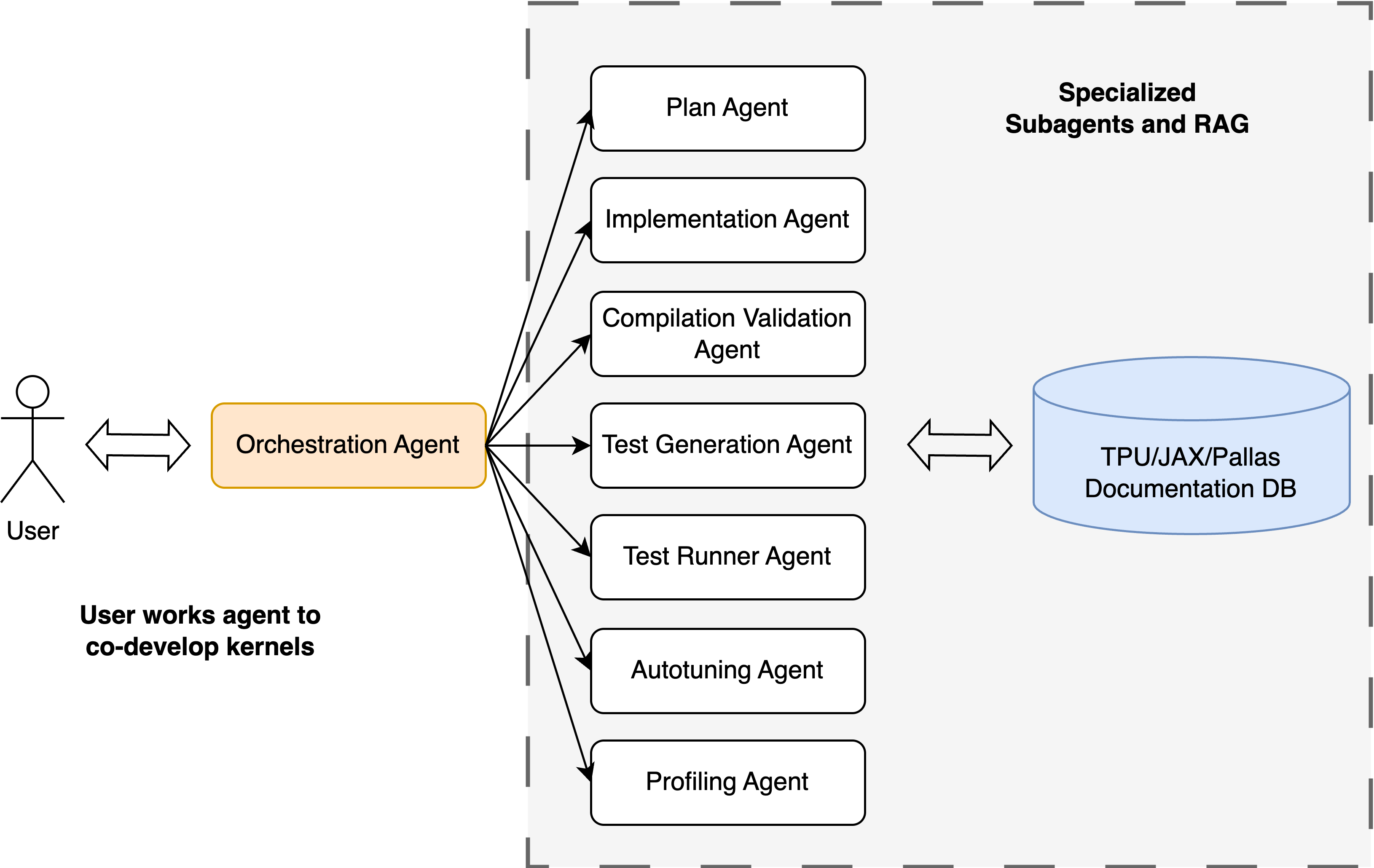}
    \caption{A schematic of the HITL Agent which demonstrates how the user has visibility into each of the sub-agents. The user has complete autonomy in choosing the sub-agents and the order of their calling.}
    \label{fig:hitl}
\end{figure}

The Human-in-the-Loop modality employs an orchestration agent acting as an interactive router. This orchestrator adheres to a "One Agent, Then Wait" paradigm~\cite{wu2023autogen}. The kernel generation process is structured into discrete phases as discussed in Section \ref{sec:subagents} (Figure~\ref{fig:hitl}) and users can seamlessly traverse the various stages. \par

After a subagent completes its designated phase, automated execution halts, and control is immediately returned to the user. This approach allows human developers to review intermediate artifacts (such as the optimization plan markdown or the code draft), provide explicit feedback, or manually correct the trajectory before the system proceeds to the next phase. This modality is particularly effective for complex kernels where expert intuition is necessary to guide the Large Language Model (LLM) through challenging design spaces.

\subsection{The Autonomous Loop (Auto) Agent}
\begin{figure}[htbp]
    \centering
    \includegraphics[width=0.9\textwidth, keepaspectratio]{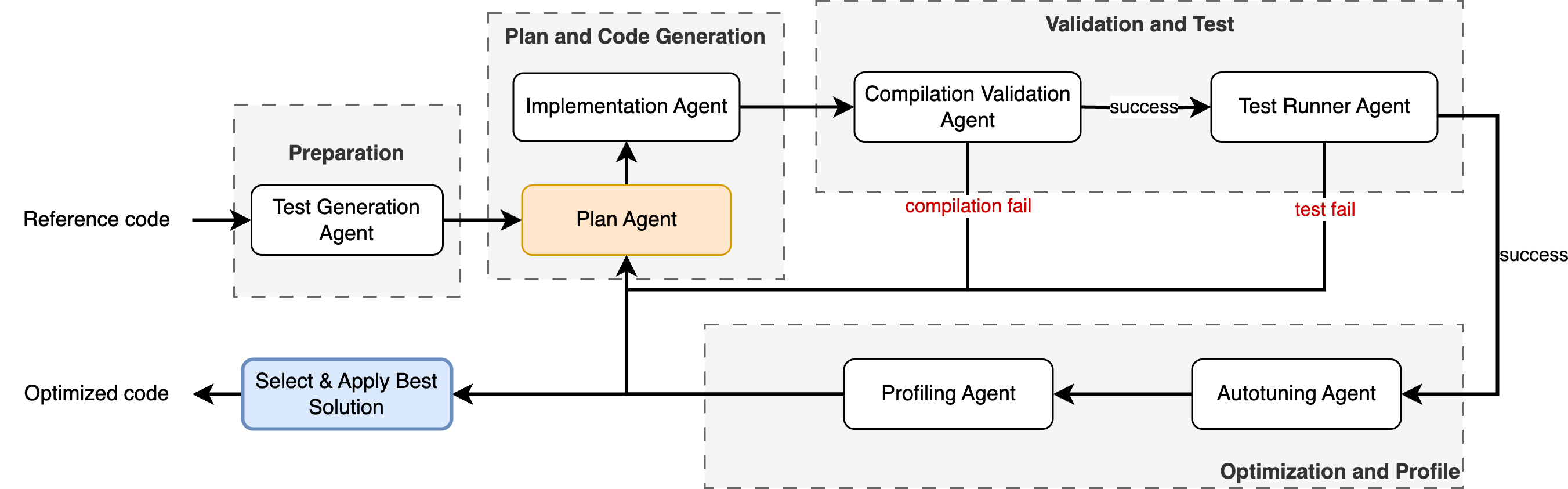}
    \caption{The closed-loop optimization workflow of the Auto Agent.}
    \label{fig:auto}
\end{figure}

For well-defined optimization tasks and large-scale benchmarking, the Auto agent chains the sub-agents into a fully automated, \textbf{closed-loop optimization pipeline} (Figure~\ref{fig:auto}). This orchestrator executes multiple iterations of the generation-evaluation cycle without human intervention, effectively performing a localized hill-climbing search.
The autonomous workflow is characterized by the following mechanisms:
\begin{itemize}
    \item \textbf{Iterative Refinement Cycle:} The orchestrator continuously loops through a strict sequence of sub-tasks: Plan Generation $\rightarrow$ Implementation $\rightarrow$ Compilation Validation $\rightarrow$ Test Execution $\rightarrow$ Autotuning $\rightarrow$ Profiling. This procedure decomposes into four distinct stages. First, in the \textit{Preparation} stage (prior to entering the loop), the system synthesizes a comprehensive test suite based on the reference code. This test suite is strictly frozen, ensuring that downstream implementation agents cannot "reward hack" or alter the validation criteria to artificially pass correctness checks. Upon entering the iterative loop, the \textit{Plan and Code Generation} stage formulates an optimization strategy and translates it into a candidate Pallas kernel. Next, the \textit{Validation and Testing} stage verifies compilability, numerical equivalence, and baseline performance against the frozen test suite; passing this stage guarantees the kernel is functionally sound. Finally, the \textit{Optimization and Profiling} stage systematically autotunes the valid kernel across various tiling configurations and captures hardware traces to identify bottlenecks. Crucially, empirical profiling feedback from this final stage is fed directly into the planning phase of the subsequent iteration, driving continuous, data-driven improvement.
    \item \textbf{Robust Failure Handling:} If a critical failure occurs---such as exceeding the maximum retry limit for compilation fixes, or failing numerical correctness checks---the pipeline immediately short-circuits. The error context is preserved, and the orchestrator loops back to the planning phase to devise an alternative strategy.
    \item \textbf{Best-of-$N$ State Rollback:} To prevent performance regressions, the orchestrator maintains a snapshot history of every successful iteration, recording the kernel code, compilation status, test results, latency, and profiling summaries. At the conclusion of the pipeline, the system evaluates the history and automatically rolls back the workspace to the optimal (lowest latency) valid solution.
    \item \textbf{Orchestrator-Driven Path Management:} To guarantee deterministic execution and prevent state corruption during autonomous loops, the root agent explicitly dictates absolute file paths for all artifacts at initialization. These rigid constraints are injected into the sub-agent prompts and file-system tools, ensuring precise file I/O tracking.
\end{itemize}

\subsection{Graph-Based Autonomous Search}
\begin{figure}[htbp]
    \centering
    \includegraphics[width=0.5\textwidth, keepaspectratio]{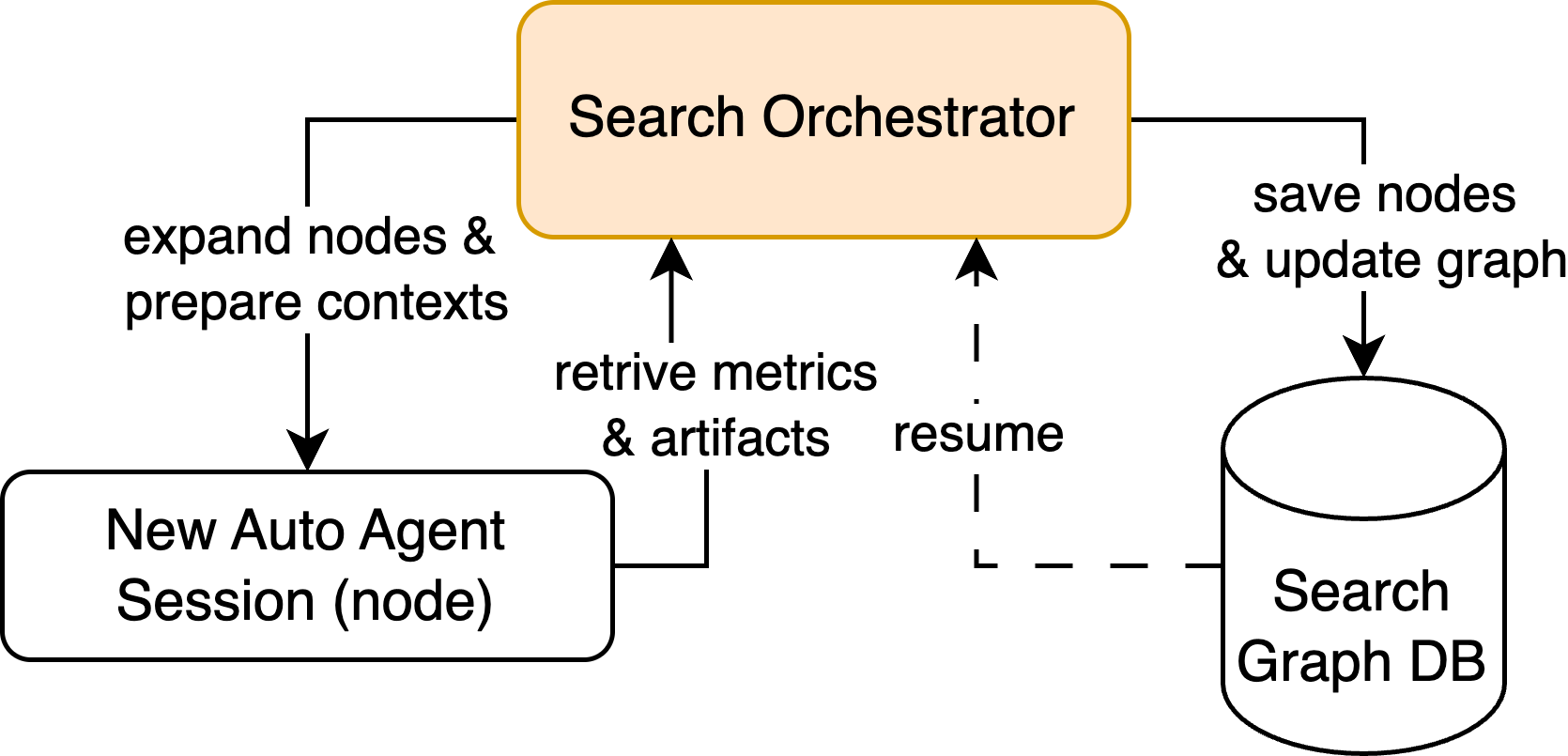}
    \caption{Design of the Graph based autonomous search. The orchestrator determines the search algorithm, where each evaluation of the search opens up a new autonomous agent session, which then retrieves and returns artifacts to the main search. A database with the saved nodes and graph is also performed at the end of each search iteration.}
    \label{fig:autosearch}
\end{figure}

While the Auto Agent performs linear, iterative improvements, it remains susceptible to local optima. To explore the optimization space more systematically, MaxKernel introduces a Graph-Based Autonomous Search (Figure~\ref{fig:autosearch}), which encapsulates the Auto Agent within a broader search algorithm. The kernel generation process is modeled as a formal search problem where the orchestrator maintains a \textit{SearchGraph}. Each node represents a specific state of the kernel produced by the auto agent worker, encompassing the source code, the optimization plan, and empirical evaluation metrics (e.g., correctness and speedup). This graph-based abstraction provides significant advantages: it enables seamless experimentation with various search heuristics, allows the system to recover gracefully from unexpected halts via the persistent graph state, and effectively mitigates LLM context window overflow by isolating each node expansion into a separate agent session.

The general search process proceeds as follows:
\begin{itemize}
    \item \textbf{Node Selection and Expansion}: The search orchestrator selects promising candidate nodes from the frontier based on specific heuristic strategies. For example, in the beam search algorithm, we select the nodes with highest speedups in each depth.
    \item \textbf{Parallel Worker Execution}: For each selected node, the orchestrator dispatches expansion tasks to distributed workers. Each worker instantiates the Auto Agent to independently apply a new optimization strategy, compile, test, and profile the resulting kernel variations in parallel.
    \item \textbf{State Update}: The results from the workers are returned to the orchestrator as new nodes. The nodes' information are incorporated in to the global graph, and the state of the search (such as the best node so far and the search frontiers) are updated.
    \item \textbf{Convergence}: This parallel, branching exploration continues until termination criteria are met (e.g., maximum depth reached or a target latency is achieved), ultimately returning the optimal node from the graph.
\end{itemize}

By orchestrating the Auto Agent across a search graph, MaxKernel scales beyond simple linear exploration, allowing it to autonomously and effectively navigate the complex trade-offs inherent in hardware-level kernel optimization.

Currently, MaxKernel implements two primary search algorithms to navigate the optimization space. \textbf{Parallel Search} acts as an unconstrained exploration mechanism; it concurrently executes multiple independent optimization trajectories from a given baseline. Because these paths do not compete, each worker is allocated a larger iteration budget, granting the agent a deep, uninterrupted horizon to continuously debug and mature complex code changes. On the other hand, \textbf{Beam Search} acts as a highly competitive exploration mechanism, maintaining a restricted frontier of the top-$k$ most promising candidate kernels. To efficiently expand this frontier across multiple depths, Beam Search restricts each candidate to a smaller iteration budget before aggressively pruning under-performing trajectories and branching only the most immediately viable strategies.

\section{Experiments and Results}

We systematically evaluate the MaxKernel system on \textbf{JaxBench}~\cite{tschand2026jaxbench}, a curated suite of 50 diverse hardware-accelerated workloads. This benchmark comprises 17 widely used operators adapted from popular LLM architectures (e.g., complex attention mechanisms) and 33 fused operators adapted from the KernelBench dataset. Spanning attention variants, dense and sparse linear algebra, complex loss functions, and highly fused operations, JaxBench provides a rigorous testbed to evaluate MaxKernel's capability for end-to-end kernel generation and hardware-specific optimization. Beyond this standard benchmark, we also demonstrate MaxKernel's robust capability to accelerate state-of-the-art (SOTA) open-source (OSS) kernels from recent architectures. By automatically generating highly efficient Pallas implementations for complex real-world workloads—such as Multi-Head Latent Attention (MLA), Qwen3-Next Gated DeltaNet, and DeepSeek-V4 Sparse Attention—MaxKernel achieves significant improvements in latency and throughput over heavily optimized JAX and human-authored baselines.

\subsection{Evaluation Metrics and Methods}
To assess both the robustness and the optimization quality of the generated kernels, we report four primary metrics: (1) The \textit{compilation rate} measures the percentage of generated kernels that successfully compile without target-specific hardware errors (e.g., VMEM allocation failures, etc). (2) The \textit{correctness rate} measures the percentage of compiled kernels that yield numerically equivalent outputs compared to the unoptimized JAX reference implementation over a suite of test inputs. To measure aggregate performance, we report the (3) \textit{geometric mean speedup} across all tasks relative to the standard XLA compiler baseline (with performance regressions floored at 1.0x). Finally, we use the (4) \textit{fast-$p$ fraction} ($\text{fast}_p$), which quantifies the proportion of tasks that both achieve strict functional correctness and exceed a specific speedup threshold $p$:
\[
\text{fast}_p = \frac{1}{N}\sum_{i=1}^N \mathbbm{1}(\text{correct}_i \wedge \text{speedup}_i > p)
\]

All empirical evaluations are conducted on TPU v6e hardware using MaxKernel's dedicated evaluation harness. For each generated kernel, the framework dynamically synthesizes an isolated testing environment that validates functional correctness against the unoptimized reference implementation. Numerical equivalence is checked using \texttt{jnp.allclose} with both absolute and relative tolerances set to $10^{-2}$ ($\text{atol} = \text{rtol} = 10^{-2}$) for most cases. To safely account for standard \texttt{bf16} variance in custom hardware-accelerated math, certain tasks' tolerance are relaxed to at most $10^{-1}$. For a complete list of the tolerance we set for each problem, please refer to the appendix \ref{sec:tol}. To measure true hardware performance, the harness integrates directly with \textit{XProf} profiling tool. This allows the system to capture precise on-device execution times while explicitly filtering out host-side JAX dispatch and compilation overheads. By relying exclusively on these strict, XProf-backed measurements on TPU v6e, we guarantee that all reported speedups reflect authentic hardware-level acceleration rather than artifactual host or framework variances.

We compare the following four generation methods:
\begin{itemize}
    \item \textbf{Best-of-$N$ (Zero-Shot):} Independent zero-shot completions conditioned on the JAX reference. The fastest correct sample out of $N$ is selected (in our experiments, we set $N=100$).
    \item \textbf{MaxKernel Auto:} A single-trajectory iterative refinement agent restricted to 5 iterations. To account for generation variance, we execute 5 independent runs per workload and report the median performance with its upper and lower bounds.
    \item \textbf{MaxKernel Parallel:} Executed via 5 concurrent MaxKernel Auto trajectories (5 iterations each). Instead of the median, it selects the single fastest correct kernel across all runs, highlighting the upper bound of unguided parallel scaling.
    \item \textbf{MaxKernel Beam:} A structured, top-$k$ guided graph search configured with a beam width of 3, a maximum depth of 3, and 2 expansion branches per node. To limit computational cost, the inner evaluation loop is strictly capped at 2 iterations per node.
\end{itemize}
\subsection{End-to-End Kernel Performance}
The performance of the four methods on JaxBench are summarized in Table \ref{tab:performance_metrics}. All large language model queries and agent interactions throughout our experiments were conducted using the Gemini 3.1 Pro model. We leave the ablations with other LLMs as future work.

\begin{table}[]
\centering
\caption{Performance Metrics Comparison. This table details the compilation rate, correctness rate, geometric mean speedup, and $\text{fast}_1$ scores across different generation strategies. MK stands for MaxKernel. For the MK Auto column, we report the median value across the five independent runs, with the lower and upper bounds shown in brackets [lower, upper].}
\label{tab:performance_metrics}
\begin{tabular}{@{}lccccc@{}}
\toprule
                       & \textbf{Best-of-N} & \textbf{MK Auto} & \textbf{MK Parallel} & \textbf{MK Beam} \\ \midrule
Compilation Rate       & 10/50          & 49/50 [48/50, 50/50]          & 50/50        & 50/50          \\
Correctness Rate       & 10/50          & 48/50 [46/50, 49/50]          & 50/50       & 50/50         \\
Geometric Mean Speedup & 1.08          & 1.39 [1.19, 1.42]           & 1.62       & 1.48          \\
$\text{fast}_1$        & 6/50           & 22/50 [18/50, 28/50]        & 34/50       & 31/50         \\ \bottomrule
\end{tabular}
\end{table}

The baseline Best-of-N approach does not have all the relevant context from real time runs, achieving a compilation and correctness rate of only $10/50$ and a geometric mean speedup near baseline ($1.08\times$). By introducing the iterative validation and optimization loop, MaxKernel's Auto agent significantly improves over the zero-shot baseline. Importantly, the Auto agent demonstrates high stability in producing compilable and numerically correct kernels. However, because a single run is susceptible to getting stuck in suboptimal local compiled states during optimization, the speedup results of Auto agent exhibit a relatively large variance, as demonstrated by its median and lower and upper bound speedup value.

This high variance motivates the need for broader search space exploration. MaxKernel Parallel search selects the best outcome of independent agent instances run in parallel, eliminating compilation failures with $50/50$ compiled and $50/50$ correct, while boosting the geometric mean speedup to $1.62\times$ and achieving a $\text{fast}_1$ score of $34/50$. Alternatively, Maxkernel Beam search provides a guided, structured search strategy that prunes suboptimal paths sequentially. Beam search achieves a highly competitive correctness rate of $50/50$ and a geometric mean speedup of $1.48\times$ with a $\text{fast}_1$ score of $31/50$.

\begin{figure}
    \centering
    \includegraphics[width=0.7\linewidth]{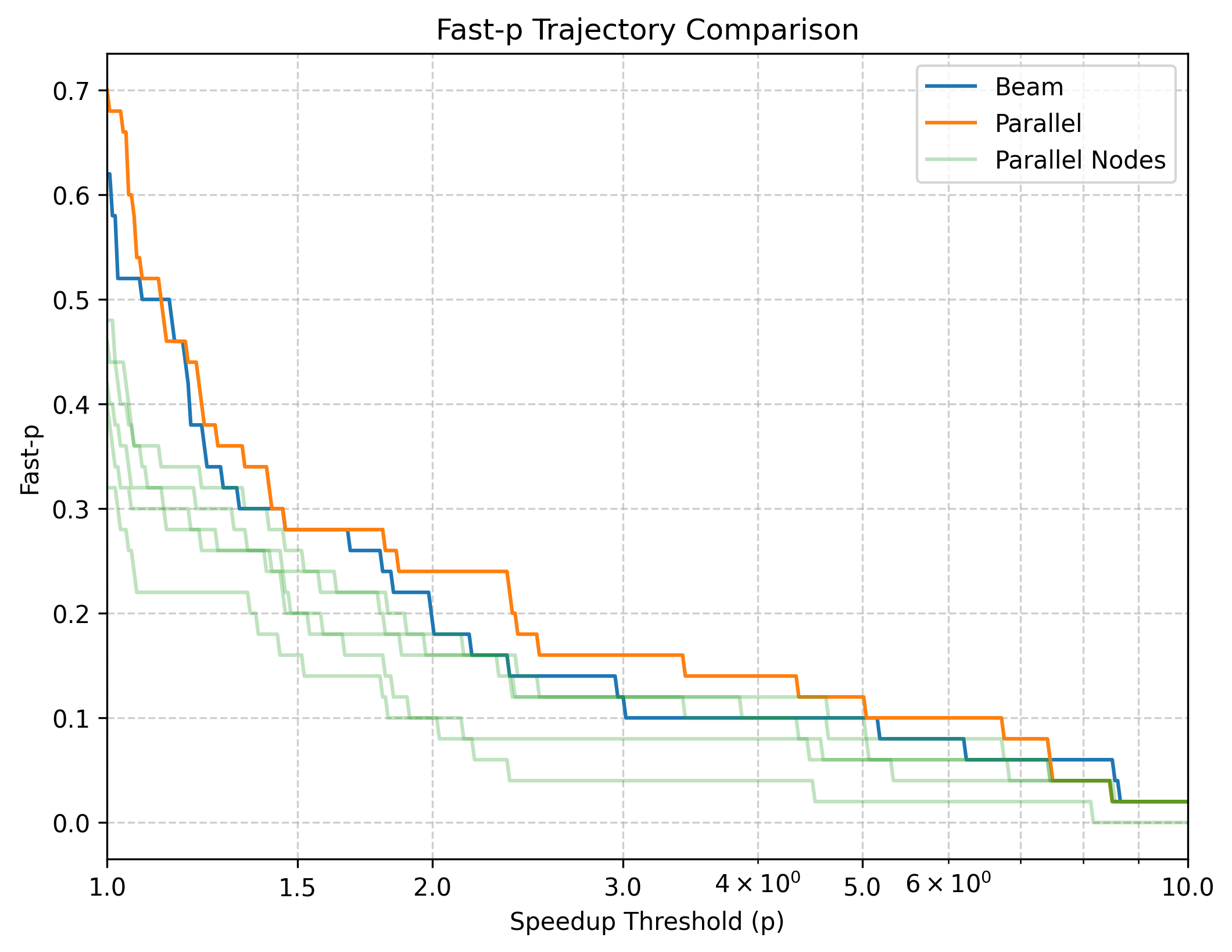}
    \caption{Comparison of the $\text{fast}_p$ metric trajectories across different methods on JaxBench for varying speedup thresholds $p$. Parallel nodes refer to the autonomous search trajectories of individual nodes.}
    \label{fig:fast_p}
\end{figure}

To show the speedup distribution of different methods, Figure \ref{fig:fast_p} compares the $\text{fast}_p$ trajectories across varying speedup thresholds p. The Best-of-N baseline is omitted from this figure due to its negligible success rate. The five individual runs of the Auto agent (shown as faded lines representing Parallel Nodes 1–5) illustrate the high performance volatility of single-pass optimization. Because parallel search is defined as the best of the independent runs of the agent, its trajectory naturally resides on top of all the individual run curves, achieving the highest success rates (e.g., $68\%$ at p=1.0 and $24\%$ at p=2.0). Importantly, the beam search trajectory also consistently sits above the individual runs for most of the speedup thresholds p. This demonstrates that structured, guided exploration successfully prunes suboptimal compilation paths and avoids local optimization minima compared to single-pass runs. 

Table \ref{tab:priority_kernels_speedups} compares the performance of our automated agentic strategies against expert-written, hand-tuned Pallas references. Remarkably, the automated agents achieve geometric mean speedups (2.69 for parallel search and 1.73 for beam search), on par with the geometric mean of the hand-tuned references (2.43). On a workload-by-workload basis, the agents achieve superior performance compared to the hand-tuned references on half of the eight workloads. For instance, on Paged Attention, the parallel search discovered a kernel that achieved a 6.74 speedup, more than double the hand-tuned reference of 2.41. The only exception where the agent demonstrate big gap against human optimized code is Ragged Paged Attention, where the human expert's code achieved a strong 4.65 speedup compared to the agent's 1.42. These results demonstrate that automated agentic compilation can consistently match or exceed human-expert optimization on complex TPU workloads, dramatically reducing developer effort.

\begin{table}
\centering
\caption{Speedups on the 8 kernels compared against JAX/XLA compile baselines and hand-tuned Pallas references. Best results for each workload is bolded.}
\label{tab:priority_kernels_speedups}
\begin{tabular}{lcccccccc}
\toprule
\textbf{Workloads} & \textbf{XLA (ms)} & \textbf{Hand-tuned}  & \textbf{MK Parallel} & \textbf{MK Beam} \\ \midrule
Flash Attention        & 15.37 & \textbf{2.47$\times$}   & 2.37$\times$ & 2.16$\times$ \\
GQA Attention          & 32.73 & 2.40$\times$   & 2.51$\times$ & \textbf{3.02$\times$} \\
MLA Attention          & 14.55 & \textbf{4.26$\times$}   & 3.87$\times$ & 1.07$\times$ \\
Sparse Attention       & 16.56  & 2.45$\times$   & \textbf{5.03$\times$} & 2.35$\times$ \\
Paged Attention        & 7.52  & 2.41$\times$   & \textbf{6.74$\times$} & 2.01$\times$ \\
Ragged Paged Attention & 15.36 & \textbf{4.65$\times$}  & 1.42$\times$ & 1.27$\times$ \\
GEMM                   & 5.41  & \textbf{1.03}$\times$  & \textbf{1.03$\times$} & 1.02$\times$ \\
Megablox GMM           & 3.11  & 1.69$\times$  & \textbf{2.39$\times$} & 1.84$\times$ \\ \midrule
Geomean (floor 1$\times$) &       & 2.43$\times$  & \textbf{2.69$\times$} & 1.73$\times$ \\ \bottomrule
\end{tabular}
\end{table}

\subsection{Parallel search vs. Beam search}
\label{subsec:linear_vs_graph}
\begin{figure}[htbp]
    \centering
    \includegraphics[width=0.75\linewidth]{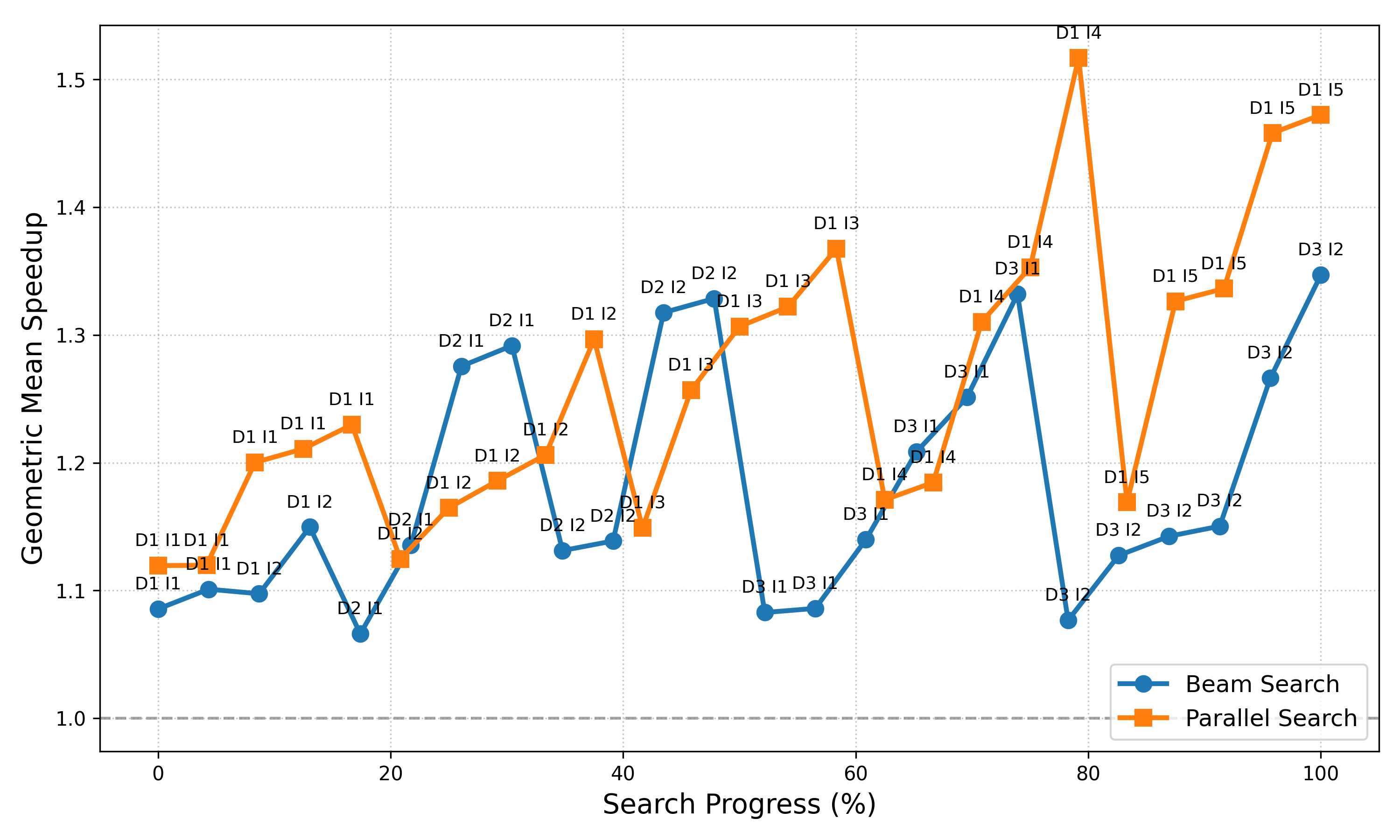}
    \caption{Progression of geometric mean speedup. Evaluated candidate nodes for Beam Search (blue) and Parallel Search (orange) are aggregated across the 50 benchmark problems. Nodes are grouped sequentially by their chronological evaluation step—denoted by Depth (D) and Iteration (I)—and sorted internally from lowest to highest performance to form the ascending segments.}
    \label{fig:iteration_progression}
\end{figure}

To evaluate the search dynamics of both strategies, we track the geometric mean speedup of the generated kernels at each topological step. Specifically, we group the evaluated kernels chronologically by their search depth and iteration, and sort the active candidate nodes within each step by their performance. Plotting these aggregated metrics sequentially yields a progression curve, where each ascending segment illustrates the performance distribution of the agent at that exact stage of the search (Figure \ref{fig:iteration_progression}).

Analyzing this progression highlights the distinct topological trade-offs between depth and breadth in automated code generation. Parallel Search, by allocating a deep refinement budget of five sequential iterations to independent trajectories, excels at stabilization. As the search advances, both the lower and upper bounds of the performance distribution rise consistently, culminating in peak geometric mean speedups exceeding 1.51$\times$. This demonstrates that long-horizon, independent refinement provides the agent sufficient opportunity to iteratively debug complex compilation errors and incrementally mature intricate memory layouts without the risk of premature pruning.

Conversely, Beam Search prioritizes exploring a wider Pareto frontier. The progression curve shows the search successfully climbing during the initial two depths before experiencing a performance plateau upon expanding to depth 3. This plateau reflects the specific, brittle nature of low-level kernel optimization in hardware-specific frameworks like Pallas. Because Beam Search enforces a shallower budget (2 iterations versus the 5 iterations in parallel search) per depth before pruning and pivoting to other strategies, the selection mechanism can occasionally penalize bold, multi-step optimizations that require longer horizons to resolve rigid memory constraints or lowering failures.

Despite this, the structural breadth of Beam Search remains advantageous depending on the nature of the optimization space. In scenarios characterized by diverse, easily compilable local minima where different algorithmic formulations, such as distinct loop orderings or mathematical reductions, yield immediate and reliable performance signals, Beam Search can be more effective. When optimizations do not require extensive, multi-step iterative debugging to compile, Beam Search's ability to broadly explore the solution space and retain a diverse set of top candidates makes it a powerful and efficient strategy for discovering optimal algorithms.

\subsection{SOTA OSS Kernel Improvement}
To evaluate the efficacy of MaxKernel, we conducted extensive benchmarking against highly optimized JAX baselines and human-authored Pallas kernels across several recent models. Our evaluation assessed improvements in latency, throughput, and compute efficiency, while also demonstrating the framework's ability to address execution robustness (Table~\ref{tab:sota_perf}).

When evaluated against human-authored Pallas baselines on the Multi-Head Latent Attention v1 architecture, MaxKernel achieved an 8.68\% reduction in latency and a 9.50\% increase in throughput. 

MaxKernel also delivered significant speedups over standard JAX implementations across a variety of architectures. For Qwen3-Next Gated DeltaNet, replacing the JAX implementation with MaxKernel-generated kernels for both forward and backward passes reduced forward pass latency by 1.63$\times$ and accelerated the overall training step by up to 4.70$\times$. On DeepSeek-V4 Sparse Attention, MaxKernel consistently outperformed the JAX baseline across all shape configurations, yielding speedups from 2.36$\times$ for small decoding shapes up to 7.85$\times$ for user-defined prefill operations. MaxKernel also effectively optimized memory-bound operations by keeping intermediate matrices fully within Vector Memory to avoid costly HBM flushes; this yielded a 1.10$\times$ speedup for the Mamba v2 State Space Duality (SSD) algorithm and a 1.66$\times$ speedup for the Compressed Sparse Attention (CSA) \texttt{StreamIndex\_topk} operation.

Beyond pure performance optimization, MaxKernel demonstrated utility in kernel debugging and robustness. We applied the framework to resolve critical crash and deadlock issues in the Ragged Page Attention v3 (RPAv3) prefill mode. MaxKernel automatically implemented protective ALU clamp instructions to handle left-padded inputs correctly, preventing negative slice sizes and ensuring DMA prefetches execute cleanly. These modifications restored crash-free execution with negligible performance overhead.

\begin{table}[h]
\centering
\caption{Performance Improvements of MaxKernel on SOTA Architectures}
\label{tab:sota_perf}
\resizebox{\columnwidth}{!}{%
\begin{tabular}{llccc}
\toprule
\textbf{Architecture} & \textbf{Workload / Metric} & \textbf{Baseline} & \textbf{MaxKernel} & \textbf{Speedup/Gain} \\
\midrule
\multirow{2}{*}{\textbf{MLA v1}} & Latency (ms) & 3.127 (Pallas) & 2.856 & +8.68\% \\
 & Throughput (TFLOPS) & 116.735 (Pallas) & 127.825 & +9.50\% \\
\midrule
\multirow{2}{*}{\textbf{Qwen3-Next GDN}} & Forward Pass (ms) & 17.09 (JAX) & 10.47 & 1.63$\times$ \\
 & Training Step (Fwd+Bwd) (ms) & 84.51 (JAX) & 17.99 & 4.70$\times$ \\
\midrule
\multirow{2}{*}{\textbf{DSv4 Sparse Attn}} & Prefill (ms) & 27.08 (JAX) & 3.449 & 7.85$\times$ \\
 & Decode (ms) & 0.047 (JAX) & 0.02 & 2.36$\times$ \\
\midrule
\textbf{Mamba v2 (SSD)} & Forward Pass (ms) & 0.211 (JAX) & 0.192 & 1.10$\times$ \\
\midrule
\textbf{StreamIndex\_topk} & Forward Pass (ms) & 0.415 (JAX) & 0.250 & 1.66$\times$ \\
\bottomrule
\end{tabular}%
}
\end{table}

\section{Conclusion}
MaxKernel demonstrates that agentic systems can successfully handle the complexities of low-level accelerator programming. In particular, we showed how a harness comprising of context synthesis, automated diagnosis, bottleneck identification via XProf profiling, and structured optimization heuristics can accelerate hillclimbing on TPUs. We demonstrated state-of-the-art performance numbers on JaxBench, reaching a geometric mean speedup of 1.62x on top of XLA optimizations. 

In the future, we want to explore other optimization algorithms too, such as evolutionary search or greedy search via the Auto Research approach. In addition, dynamic and evolving knowledge bases from past experiments could offer additional improvements. Finally, we believe stateful hybrid optimizations could help address the exploration exploitation tradeoffs across the different optimization methods.

\bibliographystyle{plain}
\bibliography{reference}

\appendix
\section{Tolerance setting for JaxBench}\label{sec:tol}
\begin{table}[htbp]
\small
\centering
\caption{Workload rtol and atol Values}
\label{tab:workload_tols}
\begin{tabular}{lcc}
\hline
\textbf{Workload} & \textbf{rtol} & \textbf{atol} \\
\hline
1p\_Flash\_Attention & 0.01 & 0.05 \\
2p\_GQA\_Attention & 0.05 & 0.05 \\
3p\_MLA\_Attention & 0.01 & 0.05 \\
4p\_Sparse\_Attention & 0.01 & 0.01 \\
5p\_Flex\_Attention & 0.01 & 0.01 \\
6p\_Paged\_Attention & 0.01 & 0.01 \\
7p\_Ragged\_Paged\_Attention & 0.01 & 0.01 \\
8p\_GEMM & 0.01 & 0.01 \\
9p\_SwiGLU\_MLP & 0.1 & 0.1 \\
10p\_Sparse\_MoE & 0.1 & 0.1 \\
11p\_Megablox\_GMM & 0.01 & 0.01 \\
12p\_RMSNorm & 0.01 & 0.01 \\
13p\_Cross\_Entropy & 0.01 & 0.01 \\
14p\_Ragged\_Dot & 0.01 & 0.01 \\
15p\_RetNet\_Retention & 0.01 & 0.01 \\
16p\_Mamba2\_SSD & 0.01 & 0.05 \\
17p\_Triangle\_Multiplication & 0.01 & 0.01 \\
18k\_Conv2D\_ReLU\_BiasAdd & 0.01 & 0.01 \\
19k\_Matmul\_Subtract\_Multiply\_ReLU & 0.01 & 0.01 \\
20k\_Gemm\_Multiply\_LeakyReLU & 0.01 & 0.01 \\
21k\_Gemm\_Divide\_Sum\_Scaling & 0.1 & 0.1 \\
22k\_Conv2d\_InstanceNorm\_Divide & 0.01 & 0.01 \\
23k\_Matmul\_Sum\_Max\_AvgPool\_LogSumExp\_LogSumExp & 0.01 & 0.01 \\
24k\_Matmul\_Scale\_ResidualAdd\_Clamp\_LogSumExp\_Mish & 0.01 & 0.01 \\
25k\_Conv3d\_GroupNorm\_Mean & 0.01 & 0.01 \\
26k\_BMM\_InstanceNorm\_Sum\_ResidualAdd\_Multiply & 0.01 & 0.01 \\
27k\_Matmul\_Mish\_Mish & 0.01 & 0.01 \\
28k\_ConvTranspose3d\_LayerNorm\_GELU\_Scaling & 0.01 & 0.05 \\
29k\_Matmul\_Swish\_Sum\_GroupNorm & 0.01 & 0.01 \\
30k\_Matmul\_Scaling\_ResidualAdd & 0.01 & 0.01 \\
31k\_Gemm\_BatchNorm\_GELU\_ReLU & 0.01 & 0.05 \\
32k\_Gemm\_Sigmoid\_LogSumExp & 0.01 & 0.01 \\
33k\_Conv3d\_Mish\_Tanh & 0.01 & 0.01 \\
34k\_Conv2d\_Activation\_BatchNorm & 0.01 & 0.1 \\
35k\_Gemm\_Scaling\_Hardtanh\_GELU & 0.01 & 0.01 \\
36k\_Matmul\_Sigmoid\_Sum & 0.01 & 0.01 \\
37k\_Matmul\_Swish\_Scaling & 0.01 & 0.01 \\
38k\_Matmul\_Dropout\_Softmax & 0.01 & 0.01 \\
39k\_Conv2d\_GELU\_GlobalAvgPool & 0.01 & 0.01 \\
40k\_Gemm\_GroupNorm\_Min\_BiasAdd & 0.01 & 0.01 \\
41k\_Gemm\_Add\_ReLU & 0.01 & 0.01 \\
42k\_Gemm\_Max\_Subtract\_GELU & 0.01 & 0.01 \\
43k\_Gemm\_BatchNorm\_Scaling\_Softmax & 0.01 & 0.01 \\
44k\_Matmul\_Divide\_GELU & 0.01 & 0.01 \\
45k\_Gemm\_GroupNorm\_Swish\_Multiply\_Swish & 0.01 & 0.05 \\
46k\_Conv2d\_GroupNorm\_Tanh\_HardSwish\_ResidualAdd\_LogSumExp & 0.01 & 0.01 \\
47k\_Matmul\_Add\_Swish\_Tanh\_GELU\_Hardtanh & 0.01 & 0.01 \\
48k\_Matmul\_BatchNorm\_BiasAdd\_Divide\_Swish & 0.01 & 0.01 \\
49k\_Matmul\_AvgPool\_GELU\_Scale\_Max & 0.01 & 0.01 \\
50k\_Matmul\_GELU\_Softmax & 0.01 & 0.01 \\
\hline
\end{tabular}
\end{table}


\newpage

\end{document}